\documentclass{article}

\usepackage{amsmath,amsfonts,bm}

\def\eqref#1{equation~\ref{#1}}

\def\1{\bm{1}}

\DeclareMathAlphabet{\mathsfit}{\encodingdefault}{\sfdefault}{m}{sl}
\SetMathAlphabet{\mathsfit}{bold}{\encodingdefault}{\sfdefault}{bx}{n}

\usepackage{microtype}
\usepackage{graphicx}
\usepackage{subcaption}
\usepackage{caption}
\usepackage{booktabs} 
\usepackage{hyperref}
\usepackage[capitalize,noabbrev]{cleveref}
\usepackage{duckuments}
\usepackage{multirow}
\usepackage{makecell}
\usepackage{enumitem}
\usepackage{colortbl}
\usepackage{xspace}
\usepackage{xcolor}
\usepackage{lipsum}
\usepackage{amssymb}
\usepackage{mathtools}
\usepackage{amsthm}
\usepackage{bbm}
\usepackage{listings}
\usepackage{wrapfig}
\usepackage{float}
\usepackage[frozencache=true,cachedir=mintedcache]{minted}

\definecolor{RowHighlight}{gray}{0.9}

\usepackage[accepted]{icml2025}

\theoremstyle{plain}

\theoremstyle{definition}

\theoremstyle{remark}

\usepackage[textsize=tiny]{todonotes}

\newcommand{\llname}{ReVISE: Learning to Refine at Test-Time via Intrinsic Self-Verification}
\newcommand{\lname}{Learning to Refine at Test-Time via Intrinsic Self-Verification\xspace}
\newcommand{\mname}{Refine via Intrinsic Self-Verification (ReVISE)\xspace}
\newcommand{\mmname}{Refine via Intrinsic Self-Verification}
\newcommand{\sname}{ReVISE\xspace}
\newcommand{\rethink}{$[\mathtt{refine}]$\xspace}
\newcommand{\eos}{$[\mathtt{eos}]$\xspace}
\newcommand{\rethinkeos}{$\{[\mathtt{eos}],[\mathtt{refine}]\}$\xspace}
\newcommand{\basestar}{STaR}
\newcommand{\baserft}{RFT}
\newcommand{\basestarplus}{STaR$^+$}

\newcommand{\placeholder}[1]{{\color{lightgray}\lipsum[1]}}

\icmltitlerunning{\lname}

\begin{document}

\twocolumn[
    \icmltitle{\llname}
    
    \icmlsetsymbol{equal}{*}
    
    \begin{icmlauthorlist}
    \icmlauthor{Hyunseok Lee}{equal,kaist}
    \icmlauthor{Seunghyuk Oh}{equal,kaist}
    \icmlauthor{Jaehyung Kim}{yonsei}
    \icmlauthor{Jinwoo Shin}{kaist}
    \icmlauthor{Jihoon Tack}{kaist}
    \end{icmlauthorlist}
    
    \icmlaffiliation{kaist}{KAIST}
    \icmlaffiliation{yonsei}{Yonsei University}
    
    \icmlkeywords{LLM, LLM reasoning, Test-time scaling}

    \icmlcorrespondingauthor{Jihoon Tack}{jihoontack@kaist.ac.kr}
    
    \vskip 0.3in
]

\printAffiliationsAndNotice{\icmlEqualContribution} 
\begin{abstract}

Self-awareness, i.e., the ability to assess and correct one's generation, is a fundamental aspect of human intelligence, making its replication in large language models (LLMs) an important yet challenging task. Previous works tackle this by employing extensive reinforcement learning or relying on large external verifiers. In this work, we propose \mmname\ (\sname), an efficient and effective framework that enables LLMs to self-correct their outputs through self-verification. The core idea of \sname is to enable LLMs to verify their reasoning processes and continually rethink reasoning trajectories based on its verification. To implement this efficiently, we introduce a structured curriculum based on preference learning. Specifically, as \sname involves two challenging tasks (i.e., self-verification and reasoning correction), we tackle each task sequentially using curriculum learning, collecting both failed and successful reasoning paths to construct preference pairs for efficient training. During inference, our approach enjoys natural test-time scaling by integrating self-verification and correction capabilities, further enhanced by our proposed confidence-aware decoding mechanism. Our experiments on various reasoning tasks demonstrate that \sname achieves efficient self-correction and significantly improves the reasoning performance of LLMs.

\end{abstract}
\section{Introduction}
\label{sec:intro}

Large language models (LLMs) have demonstrated remarkable success across diverse domains, such as coding assistants \citep{zhang2024codeagent}, search engines \citep{xiong2024search}, and personal AI assistants \citep{sajja2024artificial}, progressively advancing toward human-like logical reasoning capabilities \citep{amir2024tomeval}. However, tasks requiring rigorous System 2 thinking—such as complex reasoning \citep{jaech2024openaio1}, iterative trial-and-error \citep{song2024trial}, and dynamic planning \citep{xie2024humanplan}—remain highly challenging \citep{lowe2024system, cai2024systemmath}. A key difficulty in LLM reasoning is that errors in early steps can accumulate over time, leading to substantial inaccuracies \citep{lecun2022path}, while the models’ intrinsic ability to detect and rectify such self-generated errors—often framed as a form of self-awareness—remains insufficient. This issue is further exacerbated by the autoregressive nature of LLMs, which constrains their ability to revisit and revise prior steps \citep{bachmann2024pitfalls}.

To tackle this issue, recent approaches have emphasized verification (or correction) of LLM-generated reasoning trajectories as a crucial mechanism \citep{zhang2024mathmcts, madaan2023selfrefine}. For instance, some methods utilize external large-scale verifiers to iteratively validate outputs and trigger regeneration \citep{luo2024improve}. However, the reliance on expensive external models introduces computational inefficiencies. Alternatively, reinforcement learning (RL)-based techniques have shown promise in improving reasoning accuracy by optimizing reward signals based on ground-truth correctness, enabling self-correction \citep{kumar2024score}. However, RL is a complex and often unstable procedure \citep{mnih2015human, rafailov2023DPO}, and it does not explicitly model the verification of intermediate reasoning steps, making it difficult to assess whether a model is confident in its current trajectory or prone to deviating toward incorrect conclusions, which may limit interpretability and adaptability in complex reasoning tasks.

This raises a key question: \textit{Can LLMs be equipped with an internal mechanism to explicitly verify their own reasoning and correct potential errors based on their verification?}

\textbf{Contribution.} We propose \textbf{Re}fine \textbf{V}ia \textbf{I}ntrinsic \textbf{SE}lf-Verification (\sname)\footnote{Code available at: \href{https://github.com/seunghyukoh/revise}{github.com/seunghyukoh/revise}}, a novel and effective self-correction framework for LLM reasoning using self-verification. The core idea of \sname is to enable LLMs to assess their reasoning process and refine reasoning trajectories based on self-verification. Specifically, we introduce a special token, which outputs whether to stop the generation or revise the reasoning trajectory. To train the model to utilize this token effectively, we design a two-stage curriculum to simplify the learning of two challenging tasks—self-verification and self-correction—by breaking them into separate training stages. Here, both stages employ preference learning, allowing the model to learn these tasks efficiently without heavy computational overhead. In the first stage, we collect pairs of correct and incorrect reasoning trajectories (i.e., positive and negative samples for preference learning) based on output correctness to develop the model’s self-verification ability. In the second stage, we generate new preference pairs for self-correction by constructing positive samples where a correct reasoning path follows an incorrect one, and negative samples where an incorrect reasoning path follows a correct one.

Furthermore, we introduce an inference-time scaling strategy for \sname that leverages self-verification to enhance performance. First, as \sname inherently verifies and refines reasoning paths when it detects incorrect outputs, it naturally benefits from increased test-time computation. Additionally, we propose a novel test-time sampling scheme that incorporates self-verification confidence (i.e., the confidence in deciding whether to terminate generation). Specifically, we integrate this confidence into existing test-time sampling methods by adjusting the sampling score based on the predicted confidence, leading to more reliable output.

We demonstrated the effectiveness of \sname through evaluations on multiple reasoning datasets across mathematical and coding domains. Notably, \sname enhances reasoning performance beyond prior methods, improving accuracy from 27.1$\to$31.1\% on GSM8K (Maj@3) \citep{cobbe2021gsm8k} with Llama3 1B \citep{dubey2024llama} and from 33.2$\to$36.0\% on MATH (Maj@3) \citep{hendrycks2021math} with Llama3 8B. Furthermore, our experimental results show that \sname consistently improves accuracy without relying on external feedback mechanisms, which often degrade performance on complex reasoning tasks. For instance, unlike approaches such as Refine \citep{madaan2023selfrefine}, which struggle when combined with existing models on complex tasks, \sname achieves these gains purely through self-verification and self-correction. Finally, we show that the proposed sampling scheme is more efficient than other sampling strategies when applied to models trained with \sname, further enhancing the performance.

\section{Related Work}
\label{sec:related}

\textbf{LLM reasoning.} LLMs have made significant progress in reasoning through techniques such as Chain-of-Thought (CoT) prompting, fine-tuning, and self-improvement. CoT prompting, introduced by~\citep{wei2022emergent} and expanded by~\citep{kojima2022large} enables models to break down complex problems into intermediate steps, improving performance and interpretability. Structured reasoning methods, including self-consistency \citep{wang2022self} and Tree-of-Thought (ToT)~\citep{yao2024tree}, enhance multi-step problem-solving by exploring various reasoning paths. \citet{huang2022large} have demonstrated self-improvement through iterative feedback, refining their outputs over time. Ensuring the reliability of reasoning approaches such as Reflexion~\citep{shinn2024reflexion} and Self-Refine~\citep{madaan2023selfrefine} introduce iterative feedback loops, while verification techniques like step-by-step validation~\citep{lightman2023let} help maintain consistency and reduce errors. Unlike prior approaches, \sname learns self-verification during training, reducing train-test discrepancy and enabling more natural verification at inference.

\textbf{Test-time scaling for LLMs.} Recent works explored that scaling test-time computation, such as best-of-N sampling, can be even better than scaling train-time computation for performance \citep{snell2024testcompute}. Specifically, test-time scaling strategies improve LLM performance by generating numerous candidate outputs and selecting the best. To enhance decision-making, external verifiers are often employed to evaluate and refine these outputs \citep{liang2024improving}. Moreover, \citet{kumar2024score, qu2024rise} applied extensive reinforcement learning to overcome the efficiencies and dependence on the verifier's performance. In safety research, backtracking methods have introduced reset tokens to correct unsafe responses \citep{zhang2024backtracking}. While they focus on reducing the likelihood of unsafe outputs with limited second attempts to refuse answers, our approach targets complex reasoning tasks enabled by self-correction through an explicit verification process and two-stage curricula.

\textbf{Self-improvement for LLMs.} Self-training methods enable LLMs to refine themselves using their own outputs. Supervised fine-tuning (SFT) \citep{brown2020lmfewshot} trains on human-annotated data but lacks self-correction \citep{huang2023large}. Rejection fine-tuning (RFT) \citep{yuan2023scaling} improves robustness by filtering low-quality responses but discards useful learning signals. STaR \citep{zelikman2022star} iteratively fine-tunes models on self-generated solutions but struggles with compounding errors due to the absence of explicit verification. V-STaR \citep{hosseini2024vstar} extends STaR by jointly training a verifier alongside the generator, leveraging both correct and incorrect responses to improve self-assessment, though it still depends on large-scale self-generated data. However, discovering high-quality solutions remains a challenge, as \citep{luong2024reft} shows that RL-based fine-tuning is ineffective without supervised initialization. \citet{kim2024speculative} explore using a stronger LLM to refine incorrect rationales from a smaller model, though \citet{huang2024correctyet} argue that LLMs struggle with self-correction. Our approach integrates both generation and verification, leveraging correct and incorrect responses for more effective self-improvement.

\begin{figure*}[t]
  \includegraphics[width=\textwidth,height=8cm]{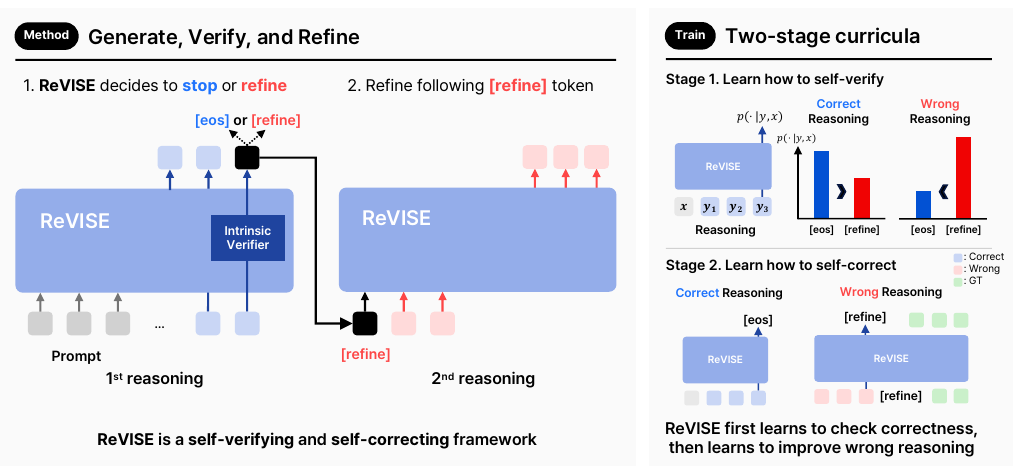}
  \caption{\textbf{Overview of \sname. Left:} \sname is a self-verifying and self-correcting reasoning framework. It first generates an initial answer, verifies its correctness, and decides whether to stop or refine. If the model generates the \rethink token, it refines the initial reasoning. \textbf{Right: } The structured curriculum-based training pipeline of \sname. In the first stage, the model learns self-verification by selecting between \eos and \rethink. In the second stage, it learns to correct reasoning mistakes using golden data.} \label{figure:method_overview}
\end{figure*}
\section{\lname} \label{sec:method}

In this section, we present \mname, an LLM reasoning framework that self-verifies and refines the reasoning trajectory based on the verification. We first introduce the problem of interest and a special token coined \rethink, which is used for refining the LLM's generation (in Section \ref{sec:problem_setup}). Then, we present the core training method, namely the two-stage curricula (in Section \ref{sec:method_rethink}) and the test-time inference strategy (in Section \ref{sec:method_scale}). The overview of \sname is depicted in Figure \ref{figure:method_overview}.

\subsection{Problem setup: Learning to Verify and Refine} 
\label{sec:problem_setup}

We describe the problem setup of our interest, i.e., self-verification and refinement. Given an input $x$, the initial output $y_\mathtt{init}$ is sampled from the LLM $\mathcal{M}$, i.e., $y_\mathtt{init} \sim \mathcal{M}(\cdot|x)$, where the reasoning path is included in $y_\mathtt{init}$. The goal is to train an LLM that verifies the correctness of $y_\mathtt{init}$ and decides whether to terminate generation or continue generating by refining its reasoning. To this end, we introduce a special token \rethink that determines whether to proceed with refinement. Specifically, given $y_\mathtt{init}$, the model verifies its correctness by predicting $v \sim \mathcal{M}(\cdot|y_\mathtt{init},x)$, where $v \in$ \rethinkeos, allowing it to either terminate generation by predicting \eos or continue generating by refining its reasoning by outputting \rethink. If refinement is needed, the model generates a revised response $y_\text{refined} \sim \mathcal{M}(\cdot|\mathtt{[refine]},y_\mathtt{init},x)$, completing the correction cycle. Note that this modeling has distinct advantages as one can access the model's verification confidence of $v$.

\subsection{\sname: \mmname}
\label{sec:method_rethink}

We first describe our core training pipeline of \sname, namely the structured curriculum based on online preference learning.  As \sname involves two challenging tasks (i.e., self-verification and refinement), we propose two-stage curricula. In the first stage, we train the LLM to intrinsically self-verify its generation by predicting the \eos or \rethink tokens. Then, at the second stage, we continually train this LLM to correct the generation when the output reasoning is wrong. For efficient and stable training, we employ preference optimization (i.e., learning from preference-based positive and negative pairs) based on our proposed preference data collection strategy. This allows us to perform structured preference learning without relying on reinforcement learning (RL), which can be computationally extensive and unstable \citep{rafailov2023DPO}.

\textbf{Stage 1: Learning to verify self-generations.} 
Given an initial LLM $\mathcal{M}_0$ and a supervised fine-tuning dataset $\mathcal{D}=\{(x_i, y_i)\}_{i}$ consisting of input-label pairs (including reasoning traces), our goal is to construct preference pairs for training $\mathcal{M}_0$. Specifically, for each input $x$, we generate a positive output $y^{+}$ and a negative output $y^{-}$.
To achieve this, we first sample multiple responses from $\mathcal{M}_0$. This allows us to obtain both correct reasoning outputs $y_{\mathtt{correct}}$ and incorrect ones $y_{\mathtt{wrong}}$, which are identified using the ground-truth answer $y$. Using these outputs, we construct a preference dataset by distinguishing two cases: (i) when the model generates the correct answer $y_{\mathtt{correct}}$, predicting \eos is preferred over \rethink, and (ii) vice versa for incorrect answers.
Concretely, given an input $x$ with its correct reasoning output $y_{\mathtt{correct}}$ and an incorrect output $y_{\mathtt{wrong}}$, we define the preference triplets $(x, y^+, y^-)$ as:
\[
\begin{cases}
\big(x, \hat{y}\oplus[\mathtt{eos}], \hat{y}\oplus[\mathtt{refine}]\big), & \text{if } \hat{y}=y_{\mathtt{correct}}\\
\big(x\oplus\hat{y}, [\mathtt{refine}], [\mathtt{eos}]\big), & \text{if } \hat{y}=y_{\mathtt{wrong}}\\
\end{cases} 
\]
where $\oplus$ is the concatenation operator.
Based on the proposed collection strategy, we generate a preference dataset $\mathcal{D}_\mathtt{verify}$ for training the intrinsic verification of the LLM. 
To this end, we jointly optimize the supervised fine-tuning loss with the direct preference optimization (DPO; \citealp{rafailov2023DPO}) loss. Specifically, for a given preference dataset $\mathcal{D}$, the SFT and DPO preference losses are defined as:
\begin{align*}
    \mathcal{L}_{\mathtt{SFT}}(\mathcal{D}):=-
    &\mathbb{E}_{(x,y^{+})\sim\mathcal{D}}\log\mathcal{M}(y^{+}|x)\\
    \mathcal{L}_{\mathtt{Pref}}(\mathcal{D}) := - 
    &\mathbb{E}_{(x,y^{+},y^{-})\sim \mathcal{D}} \Big[ \sigma\big( r(x,y^{+}) - r(x,y^{-})\big)\Big] 
    \\
    &\text{where} ~~~~~r(x,y)= \beta\log\frac{\mathcal{M}(y \mid x)}{\mathcal{M}_{0}(y \mid x)},
\end{align*}
where $\beta\in\mathbb{R}^{+}$ is hyper-parameter controlling proximity to the base model $\mathcal{M}_{0}$ and $\sigma$ is the logistic function. It is worth noting that SFT loss only focuses on minimizing the negative log-likelihood of the positive output, i.e., enforcing the model to predict the correct reasoning and answer.

Then, our training objective for self-verification is as:
\begin{equation}
    \mathcal{L}_{\mathtt{verify}} := \mathcal{L}_{\mathtt{SFT}}(\mathcal{D}_{\mathtt{verify}}) + \lambda ~ \mathcal{L}_{\mathtt{Pref}}(\mathcal{D}_{\mathtt{verify}}) 
\end{equation}
where $\lambda\in\mathbb{R}^{+}$ is a loss balancing hyperparameter. Here, we denote the initial model $\mathcal{M}_0$ trained with $\mathcal{L}_{\mathtt{verify}}$ as $\mathcal{M}_1$, which is the output model of first curricula. 

\textbf{Stage 2: Learning to correct self-generations.} We now describe how to train \sname to acquire another core ability: self-correction. Similar to self-verification, we perform preference learning using the same loss function. To this end, we aim to construct a new preference dataset, denoted as $\mathcal{D}_{\mathtt{correct}}$. The core idea consists of two main components. First, the curriculum learning: we utilize outputs generated by the model $\mathcal{M}_1$ and initialize stage 2 training from $\mathcal{M}_1$. Second, to learn how to correct incorrect outputs, we repurpose the wrong reasoning paths $y_{\mathtt{wrong}}$ used in stage 1 to construct the dataset.

Concretely, we consider two possible cases: whether the initial response is correct $y_{\mathtt{correct}}$ or incorrect $y_{\mathtt{wrong}}$. If the initial response is correct $y_{\mathtt{correct}}$, we construct preference data as same as stage 1, i.e., discouraging the generation of \rethink and encouraging \eos. The key case is when the initial response is incorrect $y_{\mathtt{wrong}}$. In this case, we need to have a positive preference sample that refines the incorrect reasoning $y_{\mathtt{wrong}}$ with the correct reasoning. To achieve this, we concatenate the ground-truth label $y$ to the response. Formally, the preference pairs are defined as:
\[
\begin{cases}
\big(x, \hat{y}\oplus[\mathtt{eos}], \hat{y}\oplus[\mathtt{refine}]\big), & \text{if } \hat{y}=y_{\mathtt{correct}}\\
\big(x \oplus \hat{y}, [\mathtt{refine}]\oplus\textcolor{blue}{y},  [\mathtt{eos}]\big), & \text{if }  \hat{y}=y_{\mathtt{wrong}}\\
\end{cases} 
\]
where $\textcolor{blue}{y}$ is the ground-truth label. Using the self-correction preference dataset $\mathcal{D}_{\mathtt{correct}}$, we train the final model $\mathcal{M}_{2}$ from $\mathcal{M}_{1}$ with the following correction loss:
\begin{equation}
    \mathcal{L}_{\mathtt{correct}} := \mathcal{L}_{\mathtt{SFT}}(\mathcal{D}_{\mathtt{correct}}) + \lambda ~ \mathcal{L}_{\mathtt{Pref}}(\mathcal{D}_{\mathtt{correct}}).
\end{equation}
It is worth noting that stage 2 explicitly defines when and how refinements should be applied, preventing overgeneration and improving response accuracy. By distinguishing between necessary and unnecessary refinements, the model ensures efficient self-correction while simulating multi-step reasoning for complex scenarios.

Furthermore, our dataset collection strategy shares similarities with recent backtracking methods in that incorrect initial generations are utilized to create negative pairs \citep{zhang2024backtracking}. We also observe that leveraging past failure trajectories aids in ultimately achieving successful reasoning. In this regard, we believe that applying \sname to safety-critical applications, akin to backtracking, is an interesting future direction, where our proposed curriculum learning and explicit self-verification stage can contribute to developing safer models.

\subsection{Verification Confidence-Aware Sampling}
\label{sec:method_scale}

We propose an inference method for models trained with \sname. The key idea is to calibrate the standard sampling-based scoring approach using the self-verification confidence. Specifically, we apply this method to majority voting, where $N$ samples are generated, and the most frequent prediction is selected. Unlike conventional approaches, our method explicitly accesses the self-verification confidence, as our model not only generates an answer but also determines its correctness by producing either an \eos or \rethink token. This allows us to directly obtain the probability associated with self-verification, enabling confidence-weighted aggregation for more reliable predictions.

\begin{table*}[htbp] 
    \caption{Accuracy (\%) for \sname (Ours) and other baselines, including Few-shot CoT, SFT, RFT, $\text{STAR}^+$ trained models. We consider two math reasoning benchmarks, GSM8K~\citep{cobbe2021gsm8k} and MATH-500~\citep{lightman2023let}. MATH-500 is a subset of the original MATH benchmark~\citep{hendrycks2021math}. Maj@K indicates that majority voting for K samples, exceptionally \sname used its own verification confidence-aware majority voting. The \textbf{bold} indicated the best result within the group.} 
    \label{tab:main}
    \vspace{0.05in}
    \small
    \centering
    \resizebox{\textwidth}{!}{
        \begin{tabular}{lcc cc cc cc}
        \toprule
        & \multicolumn{4}{c}{Llama-3.2-1B} & \multicolumn{4}{c}{Llama-3.1-8B} \\
        \cmidrule(lr){2-5}\cmidrule(lr){6-9}
        & \multicolumn{2}{c}{GSM8K} & \multicolumn{2}{c}{MATH-500} & \multicolumn{2}{c}{GSM8K} & \multicolumn{2}{c}{MATH-500}\\
        \cmidrule(lr){2-3}\cmidrule(lr){4-5}\cmidrule(lr){6-7}\cmidrule(lr){8-9}
        Methods & Maj@1 & Maj@5 & Maj@1 & Maj@5 & Maj@1 & Maj@5 & Maj@1 & Maj@5\\
        \midrule
        
        Few-shot CoT & \phantom{0}5.7  & \phantom{0}7.2 & \phantom{0}3.0&\phantom{0}3.2 & 56.7& 58.3& 23.4 &23.2  \\ 
        SFT \citep{brown2020language} & 22.1 & 26.4 & 10.4 & 11.4 & 58.2& 64.8& 27.8 & 33.2 \\
        RFT \citep{yuan2023scaling} & 26.2 & 28.6 & 12.6 & 12.8 & 58.9& 65.3& 30.8 & 35.6 \\
        STaR$^+$ \citep{zelikman2022star} & 26.2 & 29.9 & 11.4 & 13.4 & 59.2& 64.9& 30.4 & 32.8 \\
        \rowcolor{RowHighlight}\textbf{\sname (Ours)}& \textbf{28.1} & \textbf{32.8} & \textbf{13.4} & \textbf{14.8} & \textbf{61.6} & \textbf{69.2} & \textbf{33.6} & \textbf{37.6}\\
        \bottomrule
        \end{tabular}
    }
    \vspace{-.07in}
\end{table*}

Concretely, given an input $x$, we generate $N$ candidate answers $\mathcal{Y} = \{y_1, y_2, \dots, y_N\}$ from the LLM at stage 2, denoted as $\mathcal{M}$ for simplicity, where each $y_i$ is sampled as $y_i \sim \mathcal{M}(\cdot | x)$. To refine the selection process, we leverage the softmax probability of the verification (i.e., the probability of \eos token), denoted as follows:  
\[
c_i = \mathcal{M}([\mathtt{eos}] | y_i, x),
\]  
as a confidence score. Instead of selecting the most frequent prediction, we accumulate these scores by summing the confidence values of identical answers, leading to the final prediction as follows:  
\[
y^* = \arg\max_{y \in \mathcal{Y}} \sum_{i: y_i = y} c_i.
\]  
This approach calibrates the traditional majority voting method by weighting predictions based on their model-derived confidence, showing effective scaling at test time.


\section{Experiments}
\label{sec:experiments}

\begin{figure*}[t]
\begin{minipage}{\textwidth}
    \centering
    \begin{subfigure}[t]{.495\textwidth}
    \includegraphics[width=\textwidth]{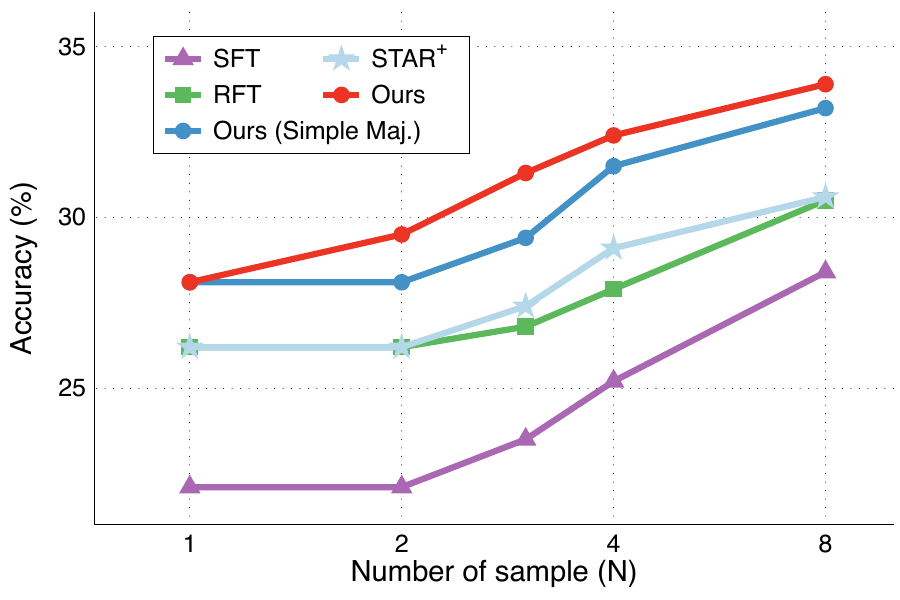} 
    \caption{Llama-3.2-1B at GSM8K}
    \end{subfigure}
    \begin{subfigure}[t]{.495\textwidth}
    \includegraphics[width=\textwidth]{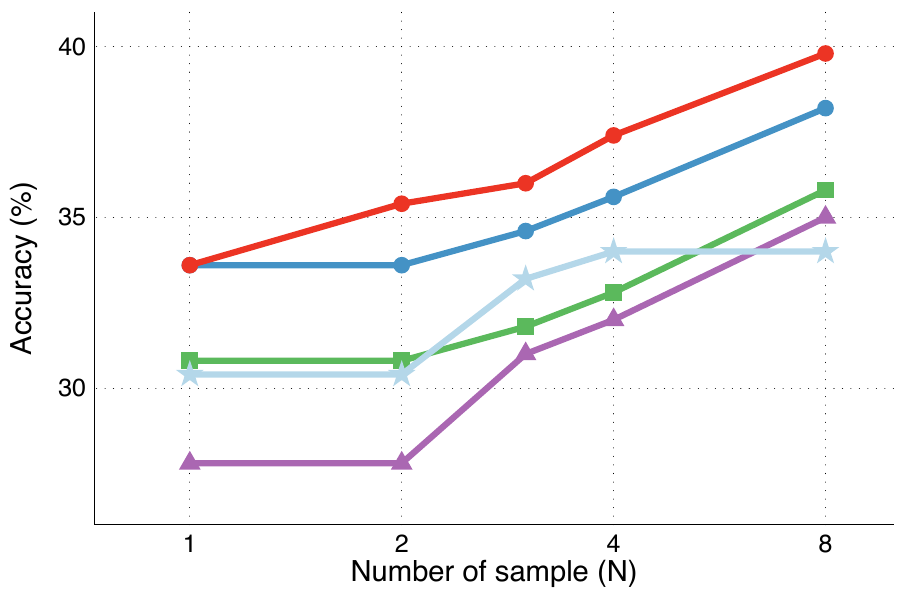} 
    \caption{Llama-3.1-8B at MATH}
    \end{subfigure}
    \vspace{-.05in}
    \caption{Test-time scaling comparison between \sname (Ours) and baselines, including SFT, RFT, STAR$^+$, and majority voting for \sname~(Ours (Simple Maj.)) at sampling sizes $N\in\{1,2,3,4,8\}$. (a) Results for Llama-3.2-1B on the GSM8K dataset. (b) Results for Llama-3.2-8B on the MATH dataset. \sname consistently outperforms baselines across all sample sizes and datasets.}
    \label{figure:test_time_scaling}
\end{minipage}
\vspace{-.1in}
\end{figure*}

We provide an empirical evaluation of \sname by investigating the following questions:
\begin{itemize}[leftmargin=*,topsep=0.0pt,itemsep=.5pt]
    \item Can \sname enhance reasoning performance? (Table \ref{tab:main})
    
    \item Does confidence-aware sampling improve the performance? (Figure \ref{figure:test_time_scaling} and Figure \ref{figure:infererence_time_scaling_comparison})
    
    \item Does/How does the proposed curriculum learning improve the performance? (Figure \ref{figure:ablation_experiments})
    
    \item Can \sname perform self-verification and -refinement? (Figure \ref{figure:verification_distribution} and Figure \ref{figure:analysis_refinement}) 
\end{itemize}

\textbf{Training setup.}
For the main experiment, we train \sname on Llama-3 models with 1B and 8B parameters, which are not instruction-tuned. We avoid using instruction-tuned models to prevent potential bias from exposure to the gold data of the tasks \citep{wang2024scpo}. For this reason, the models were first supervised fine-tuned using the labeled dataset, followed by fine-tuning with each respective method. For GSM8K \citep{cobbe2021gsm8k}, we train \sname using the original training split. For MATH \citep{hendrycks2021math}, we train \sname using a 50k subset of MetaMath \citep{yu2024metamath}, an augmented version of MATH, and use a 3k subset for the validation set, respectively. Here, MetaMath was employed to mitigate the performance degradation caused by the limited size of the original MATH.

\textbf{Baselines.} We compare our method against several baseline approaches: Supervised Fine-Tuning (SFT), \baserft \citep{yuan2023scaling}, and \basestarplus. In \baserft, fine-tuning is performed on supervised fine-tuning data $\mathcal{D}$ and correctly generated samples selected from $k$ completions for each input in the training set by a tuned model. Like \baserft,~\basestar~\citep{zelikman2022star} trains on correctly generated samples, including self-generated rationales given a hint (rationalization). However, unlike \baserft, \basestar~iteratively refines this process without relying on $\mathcal{D}$. Since both \sname and \baserft~utilize ground truth data $\mathcal{D}$, we introduce an extended version of \basestar~that incorporates SFT data as a baseline, referred to as \basestarplus. Essentially, \basestarplus~functions as a multi-iteration variant of \baserft~with rationalization. We run \basestarplus~for three iterations, sampling $k$ completions per iteration (GSM8K: $k=10$, MATH: $k=4$, GSM240K: $k=1$) with a temperature of 0.7 for both \baserft~and \basestarplus. We initialize a model for each iteration from $\mathcal{M}_{0}$ that is supervised fine-tuned with $\mathcal{D}$ at each iteration for \basestarplus to prevent overfitting.

\textbf{Evaluation setup.} We mainly report Majority Voting at K (Maj@K) as a sampling-based metric, exceptionally \sname used verification confidence-aware majority voting as described in Section \ref{sec:method_scale} (unless otherwise specified). We evaluate \sname and baselines on GSM8K \citep{cobbe2021gsm8k} and MATH-500 \citep{hendrycks2021math}, a widely used evaluation benchmark subset of MATH.

\subsection{Main Results}
We first present the main result by comparing the math problem-solving performance with other baselines. Here, we mainly compare \sname with various fine-tuning schemes that use a single network and do not use reinforcement learning. Furthermore, we demonstrate the performance of each method with simple test-time scaling methods (i.e., majority voting for baseline methods and using our verification-aware sampling for \sname). Also, we verify that \sname effectively enhances reasoning in the coding domain (i.e., MBPP~\citep{austin2021program}).

As shown in Table \ref{tab:main}, we present the math-solving performance of \sname compared to other baselines. Overall, \sname significantly and consistently outperforms all prior baseline methods. It is worth noting that for both GSM8K and MATH-500, \sname achieves the highest Maj@1, indicating that \sname is already strong without the proposed sampling scheme. For instance, \sname attains 33.6\% for Maj@1, significantly outperforming SFT (30.4\%) and few-shot CoT (23.4\%) on MATH-500 with Llama-3.1-8B. In addition, with the proposed confidence-aware majority voting, \sname marked a 4.0\% gain after refinement and consistently outperforms other baselines under five sampled answers. These results demonstrate that \sname enhances problem-solving accuracy and improves test-time scaling abilities.

\begin{table}[htbp]
    \vspace{-0.2in}
    \caption[Results on the MBPP Benchmark for \sname and Baseline Methods]{Results on the MBPP~\citep{austin2021program} benchmark for \sname and baselines trained on Llama-3.2-1B.} 
    \label{tab:main_coding}
    \vspace{0.05in}
    \small
    \centering
    \begin{tabular}{lc}
        \toprule
        Method & Pass@1 \\
        \midrule
        Few-shot CoT & 24.5\\
        SFT & 30.0\\
        RFT & 29.6\\
        STaR$^+$ & 30.7\\
        \rowcolor{RowHighlight} \textbf{\sname (Ours)} & \textbf{33.1}\\
        \bottomrule
    \end{tabular}
    \vspace{-0.05in}
\end{table}

As shown in Table \ref{tab:main_coding}, we further investigate the performance of \sname in the coding benchmark MBPP~\citep{austin2021program}.
Specifically, \sname surpasses all baseline approaches, achieving a Pass@1 score of 33.1\%, notably outperforming strong baselines such as SFT (30.0\%) and STaR$^+$ (30.7\%). These results highlight \sname's effectiveness beyond mathematical reasoning, extending its superior refinement capabilities to code-generation tasks as well. The consistent performance improvement across diverse benchmarks underscores the generalizability and robustness of the intrinsic refinement strategy employed by \sname.

\subsection{Inference Scalability of \sname} 

In this section, we evaluate the inference scalability of \sname. To this end, we visualize how the test-time scaling improves as one samples more candidates. Specifically, we conduct experiments using our method with different sample sizes $N\in\{2,3,4,8\}$ and compare with results of other baselines using majority voting. 
As shown in Figure \ref{figure:test_time_scaling}, \sname achieves significant and consistent gain in all setups. 
For instance, \sname shows a large gap with the strongest baseline RFT, showing 3.3\% of improvement in MATH-500 at $N=8$. 
Furthermore, our method even benefits under limited number of samples ($N=2$), while majority voting does not show improvement. 
This is because majority voting does not use confidence and, hence, can not benefit from small samples (e.g., if all predictions are disjoint, the majority voting does not work). 
Finally, \sname shows scalable improvements in all model configurations, ranging from relatively small 1B models to large 8B models. 
Notably, \sname achieves a significant performance gain in the 8B model, suggesting strong generalization capabilities.

\subsection{Additional Analysis and Ablation}

In this section, we provide a detailed analysis of \sname to validate the effect of each proposed component. 
Unless otherwise specified, we use a Llama-3.2-1B trained on GSM8K across all methods throughout this section.

\begin{figure*}[t] 
  
\begin{minipage}{\textwidth}
    \centering
    \begin{subfigure}[t]{.4\textwidth}
    \includegraphics[width=\textwidth]{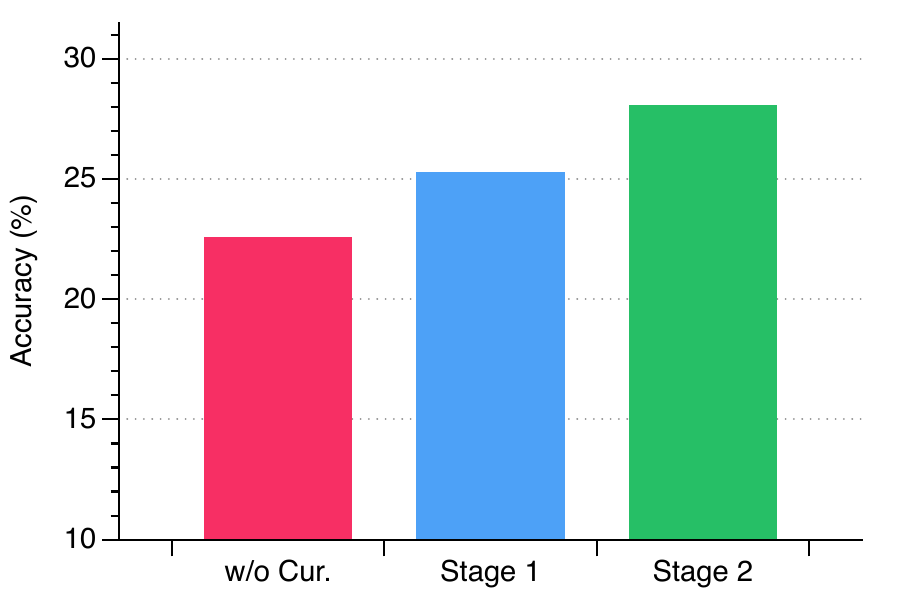} 
    \caption{Final accuracy}
    \label{subfigure:ablation_final_acc}
    \end{subfigure}
    ~~~~~~~~~~
    \begin{subfigure}[t]{.4\textwidth}
    \includegraphics[width=\textwidth]{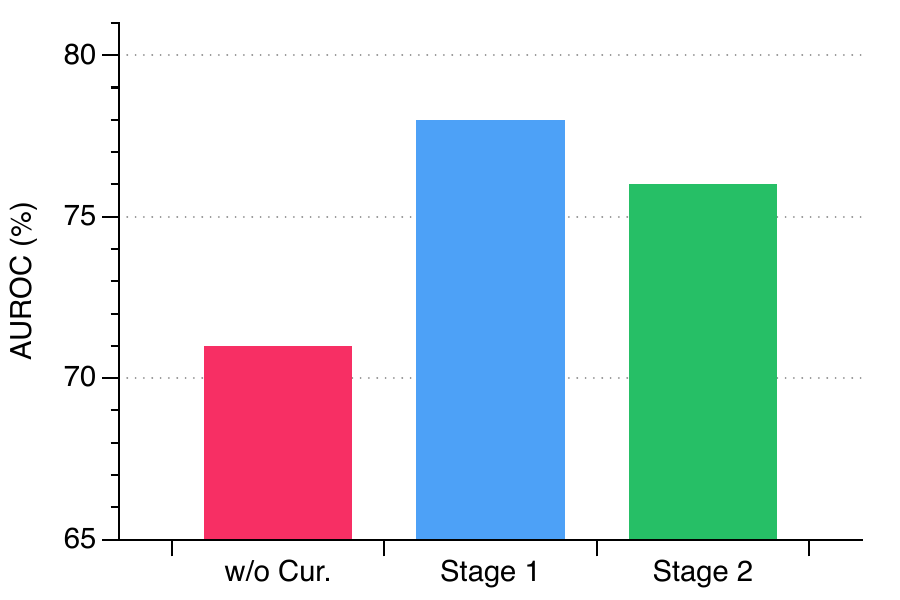} 
    \caption{Self-verification accuracy}
    \label{subfigure:ablation_self_verification_acc}
    \end{subfigure}
    \vspace{-0.06in}
    \caption{Ablation study on curriculum learning in the aspect of (a) final accuracy (\%) and (b) self-verification accuracy reported with AUROC (\%). The experiments are conducted using Llama-3.1-1B on the GSM8K dataset. The comparison includes a model trained without curriculum learning (w/o Cur.), trained for only stage 1 (Stage 1), and trained using the full two-stage curriculum learning approach (\sname) (Stage 2). (a) Accuracy improves with curriculum learning by mitigating conflicts between competing objectives during early training stages. (b) AUROC results demonstrate enhanced classification performance of corrected and incorrect responses and effective transfer from Stage 1 to the final \sname model. }
    \label{figure:ablation_experiments}
\vspace{-0.05in}
\end{minipage}
\end{figure*}

\textbf{Effectiveness of curriculum learning.}
We validate the effectiveness of the proposed curriculum learning (in Figure \ref{figure:ablation_experiments}). 
To this end, we train two types of models.
First, the model trains without curriculum by optimizing SFT $\mathcal{L}_{\mathtt{SFT}}$ and preference $\mathcal{L}_{\mathtt{Pref}}$ loss by using all preference dataset at once, i.e., $\mathcal{D}_{\mathtt{correct}}$. 
Second, we train the model only using the first verification loss, i.e., $\mathcal{D}_{\mathtt{verify}}$ (note that self-verification already enables the model to generate the answer but does not know how to correct the generation). 
As shown in Figure \ref{subfigure:ablation_final_acc}, the curriculum is indeed showing a significant improvement over no curriculum baseline (even the model has used the same preference dataset); two-stage curricula improve the performance from 22.6\% to 28.1\%.

To further investigate this phenomenon, we evaluate the self-verification accuracy of each method, which measures the model’s ability to predict whether its own output is correct. 
In Figure \ref{subfigure:ablation_self_verification_acc}, we report the verification accuracy in terms of the Area Under the Receiver Operating Characteristic Curve (AUROC) for three models. Notably, the model without curriculum learning achieves an AUROC of 71\%, while two-stage curriculum learning improves this to 76\%. 
This suggests that curriculum learning enhances self-verification, allowing the model to refine its predictions based on more reliable verification signals. 
However, we observe that training at stage 2 slightly degrades verification accuracy, indicating that the self-correction task $\mathcal{D}_{\mathtt{correct}}$ is particularly challenging and may lead to catastrophic forgetting \citep{mccloskey1989catastrophic}. 
Exploring optimization strategies that improve self-verification and self-correction without compromising overall performance remains an interesting direction for future work.

\begin{figure}[htbp]
    \centering
    \includegraphics[width=0.5\columnwidth]{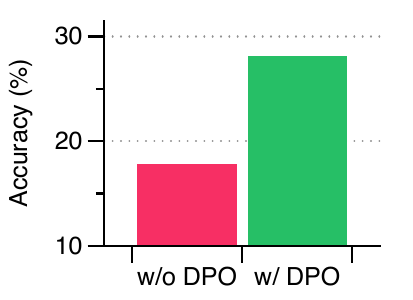}
    \vspace{-0.15in}
    \caption{Ablation study on DPO loss, evaluated on the GSM8K benchmark. Removing DPO loss significantly reduces accuracy.}
    \label{figure:ablation_dpo}
    \vspace{-0.05in}
\end{figure}
\textbf{Effectiveness of preference learning.} The role of DPO loss in \sname is to guide the model to prefer refining when the initial attempt is incorrect and terminating otherwise. Additionally, in our DPO objective, we applied SFT loss to the chosen sequence as introduced in \citet{liu2024rpoloss} which applied SFT loss to the selected sequence, $\mathcal{L}_{\mathtt{Ours}} := \mathcal{L}_{\mathtt{SFT}}(\mathcal{D}) + \lambda ~ \mathcal{L}_{\mathtt{Pref}}(\mathcal{D}) $, where $\lambda$ is a constant. 
Specifically, ablation experiments without the DPO loss—where only the SFT loss is utilized—in Figure \ref{figure:ablation_dpo} show that \sname without DPO demonstrates significantly lower performance $-10.3\%$ compared to the full-trained \sname. This indicates that the DPO loss is critical in \sname for effectively guiding the refinement process.

\begin{figure}[t]
\begin{minipage}{\columnwidth}
    \centering
    \includegraphics[width=0.8\textwidth]{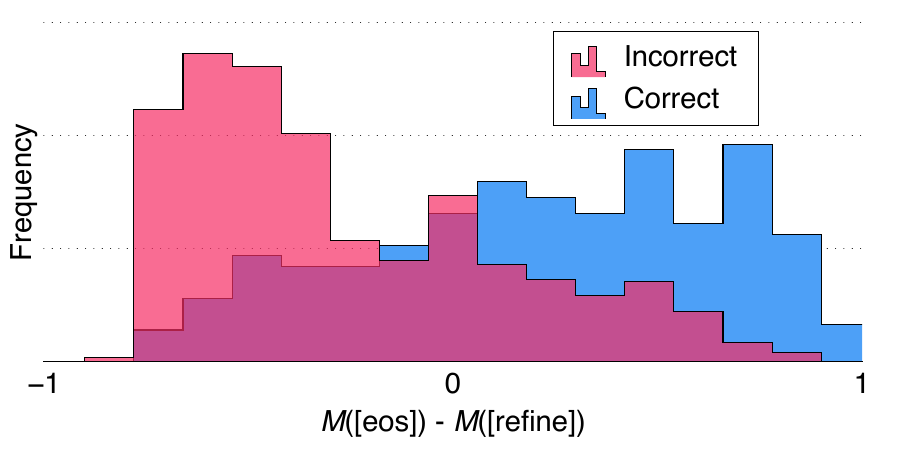} 

    \vspace{-0.1in}
    \caption{Distribution histogram of $\mathcal{M}(\text{\eos}) - \mathcal{M}(\text{\rethink})$ (ignored context $x$ for simplicity). $\mathcal{M}(\text{\eos}) - \mathcal{M}(\text{\rethink})=0$ is the threshold of \sname trigger intrinsically refine or not. Experiments are conducted using the Llama-3.2-1B model.}
    \label{figure:verification_distribution}
\end{minipage}
\vspace{-0.15in}
\end{figure}

\textbf{Analysis on the self-verification confidence of \sname.} 
We further analyze the confidence distribution in self-verification to assess whether the model's confidence is well aligned with actual correctness. To this end, we visualize the probability gap between \eos and \rethink for a given context $x$, simply defined as $\mathcal{M}([\mathtt{eos}]) - \mathcal{M}([\mathtt{refine}])$. 
As shown in Figure \ref{figure:verification_distribution}, incorrect responses tend to have lower \eos probabilities, whereas correct responses exhibit higher \eos probabilities. This demonstrates the model’s intrinsic ability to assess its own correctness. Moreover, these results suggest that confidence serves as a reliable metric for calibrating the sampling score, further validating the effectiveness of our confidence-aware sampling method.

\begin{table}[t]
    \caption{Results on the GSM8K benchmark for \sname and baselines trained on Llama-3.2-1B Instruct. Except for \sname, all methods underperform compared to the zero-shot CoT baselines.}
    \label{tab:instruction}
    \vspace{0.05in}
    \small\centering
    \begin{tabular}{lcc}
        \toprule
        Methods & GSM8K & GSM240K \\
         \midrule
        Zero-shot CoT & 48.6 & 48.6  \\ 
        SFT & 41.9 & 54.8 \\
        RFT & 44.0 & 50.9  \\
        \rowcolor{RowHighlight}\textbf{\sname (Ours)}& \textbf{52.3}  & \textbf{59.4}  \\
        
        \bottomrule
    \end{tabular}
    \vspace{-0.05in}
\end{table}
\textbf{\sname on instruction-tuned models.} 
While we have primally focused on pretrained models and initialized $\mathcal{M}_0$ with the given supervised fine-tuning dataset $\mathcal{D}$ due to the possible data contamination, we also have conducted an experiment on Llama-3.2-1B-Instruct, i.e., instruction-tuned model.
Interestingly, as shown in Table \ref{tab:instruction}, all fine-tuning methods, except for \sname, underperform the zero-shot CoT baseline when training with GSM8K. 
This outcome aligns with the widely recognized challenge that fine-tuning instruction-tuned models often leads to catastrophic forgetting, hindering their ability to learn new information effectively with a small-sized dataset.
Meanwhile, \sname remains notably resistant to this issue. We hypothesize that this advantage stems from how \sname utilizes the gold label $y$—only incorporating it as a revised second-attempt completion rather than directly fine-tuning it. In contrast, baselines such as SFT, \baserft, and \basestarplus~rely on fine-tuning the base model on $\mathcal{D}$, which becomes problematic when the target model’s performance is already strong, as it struggles to gain further improvements from $\mathcal{D}$. 

To this end, we also trained the model using GSM240K, a subset of MetaMath dataset \citep{yu2024metamath}, which expands the original data about 30-fold by rephrasing questions and answers. 
As shown in Table \ref{tab:instruction}, while training with GSM240K improved the performance of the SFT baseline, \sname still exhibited better performance. 
This result suggests that \sname can adapt to various data characteristics, even in heavily augmented settings.

\begin{figure}[t] 
    \centering
    \includegraphics[width=0.9\columnwidth]{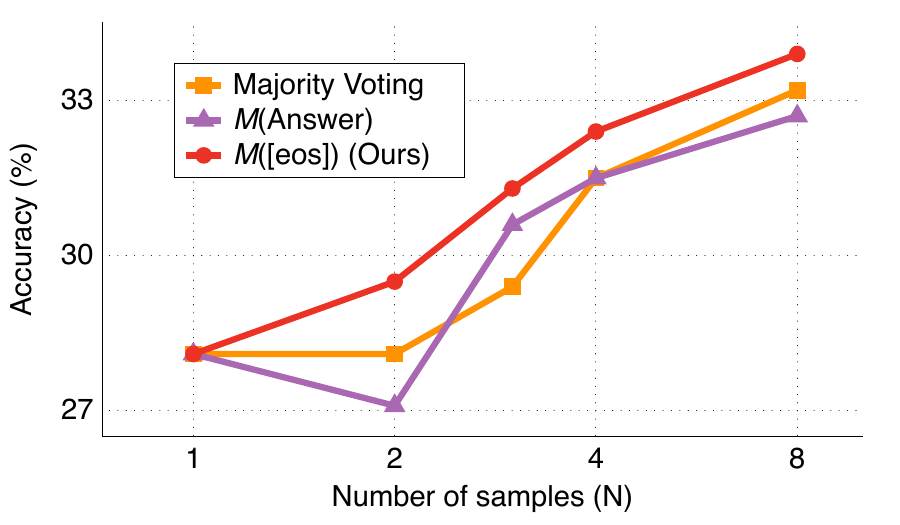}
    \vspace{-0.12in}
    \caption{Inference-time scaling comparison between \sname ($\mathcal{M}$(\eos) (Ours)) and inference metrics. For $\mathcal{M}$(Answer) and $\mathcal{M}$(\eos) (Ours), we have done weighted majority voting. \sname consistently outperforms other inference metrics, and \sname's sampling method using weighted majority voting exceeds the performance of majority voting. Experiments are conducted using the Llama-3.2-1B model.}
    \label{figure:infererence_time_scaling_comparison}
    \vspace{-0.13in}
\end{figure}
\textbf{Ablation study on confidence-aware sampling.} 
We explore the impact of different score calibrations during inference by leveraging \sname's self-verification mechanism to enable test-time compute-scalable inference strategies (see Section \ref{sec:method_scale}). 
Specifically, we compare three scoring schemes: (1) weighted majority voting using \( \mathcal{M}(\text{\eos}|x) \), (2) unweighted majority voting, and (3) scoring based on the model's predicted answer likelihood. 
These calibration methods govern both the selection of candidate answers and the evaluation of their validity. 

As shown in Figure \ref{figure:infererence_time_scaling_comparison}, \( \mathcal{M}(\text{\eos}|x) \)-based (Ours) score consistently outperforms alternatives across GSM8K benchmarks. 
For example, with eight sampled candidates, \( \mathcal{M}(\text{\eos}|x) \)-based scoring achieves an accuracy of 33.9\%, compared to 33.2\% (unweighted majority), and 32.7\% (likelihood-based). 
The trend persists across all tested sampling budgets, suggesting strong compatibility with self-verification mechanisms. 
This consistent advantage implies \( \mathcal{M}(\text{\eos}|x) \) better aligns with the model's intrinsic verification capability to distinguish correct reasoning paths. 
We carefully hypothesize that \( \mathcal{M}(\text{\eos}|x) \) acts as a latent indicator of solution correctness, as premature termination often correlates with reasoning errors.

\newcommand{\first}{\textit{First}\xspace}
\newcommand{\retry}{\textit{Retry}\xspace}

\begin{figure}[htbp]
\begin{minipage}{\columnwidth}
    \centering
    \begin{subfigure}[t]{.495\textwidth}
    \includegraphics[width=\textwidth]{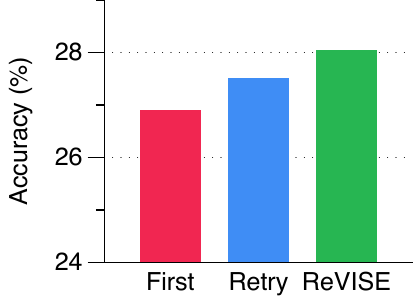} 
    \caption{GSM8K}
    \end{subfigure}
    \begin{subfigure}[t]{.495\textwidth}
    \includegraphics[width=\textwidth]{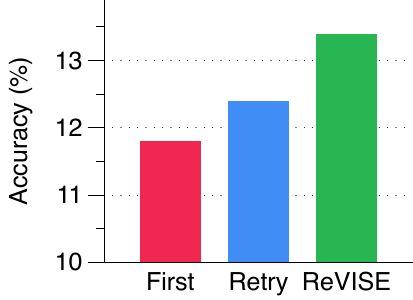} 
    \caption{MATH}
    \end{subfigure}
    \caption{Analysis of refinement capability of \sname. We compare the accuracy (\%) on GSM8K and MATH when using different decoding approaches. \first stops at \rethink, \retry re-generates responses, while \sname refines its initial reasoning. The results show that \sname improves accuracy, demonstrating its ability to refine rather than randomly re-generate responses. Experiments are conducted using the Llama-3.2-1B model.}
    \label{figure:analysis_refinement}
\end{minipage}
\vspace{-0.1in}
\end{figure}
\begin{figure*}[h]
    \centering
    \begin{subfigure}[t]{.495\textwidth}
        \centering
        \includegraphics[width=0.6\textwidth]{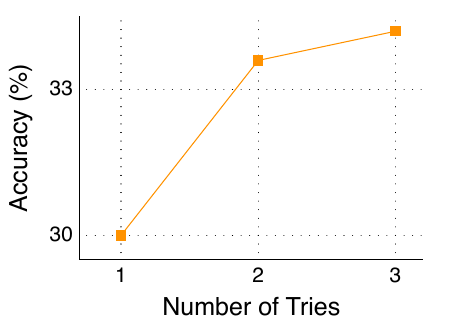} 
        \caption{Llama-3.2-8B fine-tuned in MATH and evaluated at MATH-500}
    \end{subfigure}
    \begin{subfigure}[t]{.495\textwidth}
        \centering
        \includegraphics[width=0.6\textwidth]{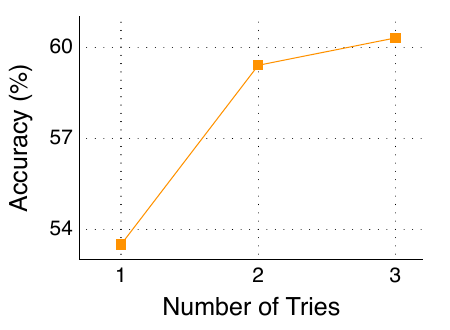} 
        \caption{Llama-3.2-1B-instruct evaluated at GSM8K}
    \end{subfigure}
    \caption{Accuracy improvements through iterative refinement. The plot shows the accuracy (\%) at GSM8K and MATH-500 of \sname across multiple rounds of iterative refinement (1, 2, and 3 tries). }
    \label{figure:iterative_refinement}
    \vspace{-0.09in}
\end{figure*}
\textbf{Analysis on the refinement.} We demonstrate that \sname refines its answers based on the initial attempt rather than randomly generating a new completion. 
To evaluate this, we compare \sname with two baselines: \first and \retry. 
\first terminates decoding at the \rethink token, while \retry generates a new completion upon encountering \rethink.
Specifically, \retry greedily decodes the first attempt, and if \rethink appears, it samples a new completion with a temperature of $0.7$ following the prompt $x$. In contrast, both \first and \sname greedily generate completions.
As shown in Figure \ref{figure:analysis_refinement}, \sname outperforms both \first and \retry. This result highlights that \sname does not generate new responses arbitrarily but instead meaningfully refines and improves upon its initial answer.

\begin{table}[t]
\vspace{-0.1in}
    \caption{Accuracy (\%) on GSM8K under transfer domain generalization. All models are trained on MATH and evaluated on GSM8K. Trained on Llama-3.2-1B and Llama-3.1-8B models.}
    \label{tab:ood}
    \vspace{0.05in}
    \small\centering
    \begin{tabular}{lcc}
        \toprule
        Model & Methods & Accuracy (\%) \\
         \midrule
        \multirow{4}{*}{Llama-3.2-1B} & SFT & 7.3 \\
        &RFT & 8.2  \\
        &STaR$^{+}$ & 8.0  \\
        &\textbf{\sname (Ours)}& \textbf{8.8}  \\
        \midrule    
        \multirow{4}{*}{Llama-3.1-8B} & SFT & 60.3 \\
        &RFT & 60.3  \\
        &STaR$^{+}$ & 58.7  \\
        &\textbf{\sname (Ours)}& \textbf{61.5}  \\
        \bottomrule
    \end{tabular}
    \vspace{-0.18in}
\end{table}
\textbf{Generalization under transfer dataset domain.} We demonstrate the generalization ablity of \sname in a transfer domain setting (see Table~\ref{tab:ood}). Specifically, we train \sname on the MATH domain and test on the GSM8K domain, using both Llama-3.2-1B and Llama-3.1-8B models. As shown in Table~\ref{tab:ood}, both model sizes significantly outperform other baselines in this out-of-distribution evaluation. For example, \sname achieves an accuracy of 8.8\% using Llama-3.2-1B, and 61.5\% using Llama-3.1-8 B. These results demonstrate that our method possesses strong domain transferability, maintaining its advantage over baselines even when evaluated on a different dataset.

\textbf{Iterative refining sequentially at test time.} Although \sname is trained to refine its output in a single pass, we explore its potential for iterative refinement. Specifically, after generating the second attempt, we append it to the original prompt $x$ and treat it as the first-attempt output. This allows the model to either output \eos~to terminate the sequence or generate a third attempt following the same process, effectively enabling multiple rounds of refinement. As shown in Figure \ref{figure:iterative_refinement}, in MATH-500, \sname trained on Llama-3.1-8B's accuracy consistently improves as the model iteratively refines its responses up to 3 times.

This observation suggests the potential for training a model explicitly designed for sequential iterative refinement to enhance the iterative refinement even more. By incorporating iterative refinement directly into the training process, the model could learn to self-correct more effectively across multiple rounds. We leave this direction as an exciting avenue for future work.
\section{Conclusion}
\label{sec:con}
In this paper, we introduce \mmname\ (\sname), a novel framework that enables Large Language Models (LLMs) to perform self-verification and self-correction during inference. Through a structured curriculum learning approach, we demonstrated how LLMs can progressively learn to verify their reasoning and improve their outputs. Our results across various reasoning benchmarks show that \sname significantly improves first-attempt accuracy while maintaining efficiency. Furthermore, the self-verification mechanism and a confidence-aware decoding strategy enhance model performance without introducing additional computational overhead.

\section*{Impact Statement}

This work advances the reasoning of Large Language Models (LLMs) by introducing \mname, a framework that enables self-verification and self-correction. This has potential applications requiring precise reasoning, such as automated tutoring and decision support systems. Specifically, \sname can benefit safety by rigorously revising the model's response in the aspect of the model.

\clearpage

\section*{Acknowledgements}

This work was supported by Institute for Information \& communications Technology Planning \& Evaluation (IITP) grant funded by the Korea government(MSIT) (No.RS-2019-II190075 Artificial Intelligence Graduate School Program(KAIST); No. RS-2024-00509279, Global AI Frontier Lab) and NIPA(National IT Industry Promotion Agency), through the Ministry of Science and ICT (Hyperscale AI flagship project).

\bibliography{ref}
\bibliographystyle{icml2025}

\newpage
\appendix
\onecolumn
\section{Experimental Details}
In this section, we describe the experimental details of Section \ref{sec:experiments}, including \sname and the baselines.

\textbf{Dataset details.}
In this section, we describe the dataset we used in training and evaluation. Also, explain how we generated the additional datasets.

\begin{itemize}[itemsep=0.05in,leftmargin=0.2in,topsep=0.02in]
    \item{\textbf{Grade School Math 8K (GSM8K).}} The GSM8K dataset \citep{cobbe2021gsm8k} consists of 8,790 high-quality grade-school math word problems. We used the provided train and test splits, ensuring consistency across all experiments. The dataset serves as a benchmark for evaluating the arithmetic and reasoning capabilities of language models.
    
    \item{\textbf{MATH.}} The MATH dataset \citep{hendrycks2021math} is a challenging collection of problems from high school mathematics competitions, covering diverse topics such as algebra, geometry, calculus, and statistics. We utilized the original train and test splits, which include approximately 12,500 problems. Due to the dataset’s complexity, it effectively evaluates the model's ability to handle higher-level mathematical reasoning.
    
    \item{\textbf{MetaMath.}} MetaMath \citep{yu2024metamath} is an augmented version of the MATH and GSM dataset, designed to address the challenges posed by the limited size of the original dataset. We selected a 50k subset of MetaMath for training and sampled 3k problems for the validation set. MetaMath includes additional examples generated using synthetic data augmentation techniques, such as problem paraphrasing and structural variations, to enhance diversity and improve generalization. This augmentation mitigates performance degradation associated with small datasets while maintaining the original problem difficulty and format.

    \item{\textbf{MBPP.}} MBPP~\citep{austin2021program} is a collection of crowd-sourced Python programming problems. Each instance consists of a natural language task description, a reference solution, and three test cases written in Python. Since \sname requires intermediate reasoning steps not provided in the original dataset, we generated them by applying Chain-of-Thought prompting to GPT-4o. For each problem, we collected 16 valid reasoning paths along with corresponding code solutions that successfully pass all test cases.

\end{itemize}

\textbf{Training details of \sname} We use AdamW optimizer with a learning rate $\mathtt{lr}\in\{10^{-4},10^{-5}\}$ with 10\% warm up and cosine decay and train it for one epoch. We trained with batch size 32 for fine-tuning and 64 for preference tuning . For the $\lambda$ constant for SFT loss, we used $\lambda=0.1$. During the training, for the data sampling phase, we sampled 10 times for each sample in GSM8K and 4 times for each sample in MATH.

\textbf{Training model details.} We mainly used the open-source Large Language Models (LLMs) from Llama-family. Specifically we used meta-llama/Llama-3.2-1B,  meta-llama/Llama-3.1-8B which are not instruction-tuned and meta-llama/Llama-3.2-1B-Instruct, which is instruction-tuned. We used the model checkpoint from huggingface library.

\textbf{Evaluation details.} Used lm-eval-harness for greedy decoding experiments and used our code to evaluate models in sampling settings. Since the output depends on the evaluation batch size, we fixed the batch size to 128 for a fair comparison.

\begin{itemize}[itemsep=0.05in,leftmargin=0.2in,topsep=0.02in]
    \item{\textbf{GSM8K, MBPP.}} Used the test split as a benchmark dataset.
    
    \item{\textbf{MATH-500.}} The MATH-500 dataset is a curated collection of 500 MATH dataset. For our experiments, we used Math-500 exclusively for evaluation purposes.
\end{itemize}

\textbf{Resource details.} For the main development we mainly use Intel(R) Xeon(R) Platinum 8480+ CPU @ 790MHz and a 8 NVIDIA H100 GPUs. Additionally, we used NVIDIA RTX4090 GPUs for evaluation.

\textbf{Baseline details.} 

\begin{itemize}[itemsep=0.05in,leftmargin=0.2in,topsep=0.02in]
    \item{\textbf{SFT}} We fine-tuned the model using a language modeling loss, exploring learning rates from \(1\text{e}^{-6}\) to \(1\text{e}^{-4}\), with epochs ranging from 1 to 3 and a batch size of 32.
    
    \item{\textbf{RFT}} We sampled ten completions for GSM8K, one for GSM240K, and four for MATH-50K. The model was trained for one epoch on the collected dataset with a fixed learning rate of \(1\text{e}^{-5}\).

    \item{\textbf{STaR$^+$}} We sampled the same number of samples as in \textbf{RFT}. The outer loop was fixed to 3 for all datasets, with one epoch per outer loop. Rationalization was performed with a hint, where the answer was provided except for the rationale, which served as the hint. The learning rate was fixed at \(1\text{e}^{-5}\).
\end{itemize}

\clearpage

\section{Additional Results}




\subsection{Comparison with DPO}
\begin{table}[h]
\vspace{-0.1in}
    \caption{Comparison between DPO and \sname. We report accuracy (\%) on GSM8K and MATH-500. The models are trained on Llama-3.2-1B. The \textbf{bold} indicates the best result within the group.}
    \label{tab:dpo}
    \vspace{0.05in}
    \small\centering
    \begin{tabular}{lcc}
        \toprule
        Method& GSM8K & MATH-500 \\
         \midrule
        DPO & 22.6 & 10.8 \\
        \rowcolor{RowHighlight}\textbf{\sname (Ours)} & \textbf{28.1} & \textbf{13.4}  \\
        \bottomrule
    \end{tabular}
    \vspace{-0.05in}
\end{table}
To further analyze the effectiveness of \sname's training framework, we include a comparison with a reinforcement learning(RL)-based baseline using Direct Preference Optimization (DPO). We trained the DPO model from the same supervised fine-tuned checkpoint, which was used in training \sname. We construct a preference pair dataset where ground truth answers are preferred over incorrect responses. As shown in Table~\ref{tab:dpo}, \sname even outperforms the DPO trained baseline, indicating 22.7\% in GSM8K and 10.8\% in MATH-500.

\subsection{Iterative Training and Comparison with Self-Correction Work.}
\begin{table}[h]
\vspace{-0.1in}
    \caption{Comparison between \sname and other baselines (i.e., Zero-shot CoT, SFT, RFT, and SCoRe~\citep{kumar2024score}). We report accuracy (\%) on MATH-500. The models are trained on Gemma-2-2B. The \textbf{bold} indicates the best result within the group.}
    \label{tab:iterate}
    \vspace{0.05in}
    \small\centering
    \begin{tabular}{lcc}
        \toprule
        Method& Accuracy (\%) & Training Efficiency \\
         \midrule
        Zero-shot CoT & 16.8 & - \\
        SFT & 17.6 & - \\
        RFT & 18.6 & - \\
        SCoRe & 23.0& x1  \\
        \rowcolor{RowHighlight}\textbf{\sname (Ours)} & 23.2 & \textbf{x30}  \\
        \rowcolor{RowHighlight}\textbf{+ iter1 (Ours)} & 24.2 & x20  \\
        \rowcolor{RowHighlight}\textbf{+ iter2 (Ours)} & \textbf{25.8} & x15  \\
        \bottomrule
    \end{tabular}
    \vspace{-0.05in}
\end{table}
We include a comparison with the self-correction baseline, SCoRe~\citep{kumar2024score}. While SCoRe relies on costly online reinforcement learning and requires extensive reasoning path generation (resulting in approximately 1.5 million generations for 3,000 steps with a batch size of 512), \sname is significantly more efficient. Specifically, \sname constructs preference pairs by generating a single reasoning path per sample, totaling only 50,000 generations for the entire dataset. This corresponds to a 30× reduction in training cost compared to SCoRe. Despite this large efficiency gap, \sname achieves higher accuracy than SCoRe on the MATH-500 benchmark using the same Gemma2-2B model, as shown in Table~\ref{tab:iterate}. Furthermore, \sname’s performance improves further with repeated training cycles. At each cycle, we re-sample reasoning path pairs with the current model and iteratively apply preference optimization to refine it further. This strategy leads to continual accuracy gains while maintaining substantial efficiency benefits. For instance, after two additional iterations, \sname achieves 25.8\% accuracy, which further increases its performance margin over SCoRe while incurring a training cost that is 15 times smaller.

These results demonstrate that \sname is not only more practical and scalable for self-correction tasks, but also able to leverage iterative refinement to reach even higher performance. All comparisons utilize SCoRe’s results from the original paper, as there is a lack of open-source code.

\subsection{Quantify the Verifying Performance.}
\begin{table}[h]
\vspace{-0.1in}
    \caption{Comparison of AUROC (\%) of verifying correctness between \sname and V-STaR~\citep{hosseini2024vstar} verifier. We report AUROC (\%) on GSM8K. The models are trained on Llama-3.2-1B.}
    \label{tab:appendix_verifier}
    \vspace{0.05in}
    \small\centering
    \begin{tabular}{lcc}
        \toprule
        Method& AUROC (\%) \\
         \midrule
        V-STaR verifier & 69.5 \\
        \rowcolor{RowHighlight}\textbf{\sname (Ours)} & \textbf{76.0} \\
        \bottomrule
    \end{tabular}
    \vspace{-0.05in}
\end{table}

We further evaluate the quality of the verifying signal produced by \sname using the area under the receiver operating characteristic curve (AUROC) and compare with V-STaR~\citep{hosseini2024vstar} verifier, while we reported \sname's calibration performance using AUROC in Section~\ref{sec:experiments} and Figure~\ref{subfigure:ablation_self_verification_acc}. As shown in Table~\ref{tab:appendix_verifier}, \sname achieves an AUROC of 76.0\%, outperforming the V-STaR verifier, which achieves 69.5\%, even though V-STaR uses a separately trained verifier. These experimental results demonstrate the effectiveness of \sname's intrinsic verifier, resulting in improved performance and enhanced test-time scalability.

\subsection{Extended Test-time Scaling Behavior Experiment.}
\begin{table}[h]
    \centering
    \caption{Test-time scaling results (Maj@K) on GSM8K with Llama-3.2-1B. We evaluate accuracy (\%) as the number of sampled generations K increases from 2 to 64. We compared \sname and other baselines (i.e., SFT, RFT, and STaR$^+$).}
    \vspace{0.05in}
    \label{tab:appendix_scaling}
    \begin{tabular}{ccccccc}
    \toprule
          & Maj@2& Maj@4& Maj@8& Maj@16&Maj@32&  Maj@64\\
    \midrule
    SFT & 20.5\% {\scriptsize$\pm$ 0.5} & 24.5\% {\scriptsize$\pm$ 0.6} & 28.2\% {\scriptsize$\pm$ 0.6} & 30.0\% {\scriptsize$\pm$ 0.4} & 31.8\% {\scriptsize$\pm$ 0.1} & 32.1\% {\scriptsize$\pm$ 0.4} \\
    RFT & 24.6\% {\scriptsize$\pm$ 0.3} & 27.5\% {\scriptsize$\pm$ 0.3} & 29.8\% {\scriptsize$\pm$ 0.4} & 30.9\% {\scriptsize$\pm$ 0.5} & 31.3\% {\scriptsize$\pm$ 0.3} & 33.2\% {\scriptsize$\pm$ 0.2} \\
    STaR$^+$ & 24.0\% {\scriptsize$\pm$ 1.1} & 27.1\% {\scriptsize$\pm$ 0.6} & 29.3\% {\scriptsize$\pm$ 0.6} & 30.4\% {\scriptsize$\pm$ 0.6} & 31.1\% {\scriptsize$\pm$ 0.5} & 31.6\% {\scriptsize$\pm$ 0.4} \\
    \rowcolor{RowHighlight} \textbf{\sname (Ours)} & \textbf{28.3\%} {\scriptsize$\pm$ 0.7} & \textbf{32.5\%} {\scriptsize$\pm$ 0.9} & \textbf{34.9\%} {\scriptsize$\pm$ 0.5} & \textbf{36.2\%} {\scriptsize$\pm$ 0.4} & \textbf{37.2\%} {\scriptsize$\pm$ 0.5} & \textbf{37.7\%} {\scriptsize$\pm$ 0.5} \\
    \bottomrule
    \end{tabular}
\end{table}

We extended our experiments on test-time scaling behavior of \sname and other baselines (i.e., SFT, RFT, and STaR$^+$) on GSM8K using the Llama-3.2-1B models. Specifically, for the baselines, we measure accuracy by majority voting over K sampled generations, where K ranges from 2 to 64 (Maj@K). For \sname, we used our own method, confidence-based weighted majority voting, for the same generation ranges. As shown in Table~\ref{tab:appendix_scaling}, \sname consistently outperforms every baseline across every sample size. Notably, \sname reaches 37.7\% accuracy at Maj@64, representing a substantial improvement over prior methods. These results demonstrate the effectiveness of \sname in leveraging increased generation diversity for robust test-time self-verification.

\clearpage

\subsection{\sname Results Example}

\subsubsection{Refinement via intrinsic self-verification example}
\begin{table}[h]
    \caption[]{An example of \sname's refinement on a GSM8K question}
    \label{tab:appendix_refine_gsm}
    \footnotesize
    \centering
    \begin{tabular}{>{\raggedright\arraybackslash}p{2.5cm}>{\raggedright\arraybackslash}p{12.5cm}}
        \toprule
        Question & Mrs. Tatiana owns a grocery store that sells different fruits and vegetables, which includes carrots. The price of carrots in the grocery store increases by 5\% of the original price every year. What would be the price of carrots after three years if it was \$120 initially? (Round to the nearest integer) Let's think step by step. Put your final answer at the end with 'The answer is: .' \\
        \midrule
        Response & The price of carrots increases by 5/100 * \$120 = \$6. So the price of carrots after one year is $120 + $6 = \$126. The price of carrots after two years is $126 + $6 = \$132. The answer is: $\boxed{132}$\linebreak
        {\color{red} \rethink}
        The price of carrots increases by 5/100*\$120 = \$6 per year.The price after one year is \$120 + \$6 = \$126 The price after two years is \$126 + \$6 = \$132.  The price after three years is \$132 + \$6 = \$138. The answer is: $\boxed{138}$ \\
        \midrule
        Target Answer & 138 \\
        \bottomrule
    \end{tabular}
\end{table}

\begin{table}[h]
    \caption[]{An example of \sname's refinement on a MATH question}
    \label{tab:appendix_refine_gsm}
    \footnotesize
    \centering
    \begin{tabular}{>{\raggedright\arraybackslash}p{2.5cm}>{\raggedright\arraybackslash}p{12.5cm}}
        \toprule
        Question & Ten treeks weigh as much as three squigs and one goolee. Two treeks and one goolee are equal in weight to one squig. The combined weight of how many treeks equals the weight of one squig? Let's think step by step. Put your final answer at the end with 'The answer is: .'\\
        \midrule
        Response & We can set up a system of equations to solve this problem.
        Let $t$ be the weight of one treek, $s$ be the weight of one squig, and $g$ be the weight of one goolee.
        From the given information, we can write the following equations:
        $10t = 3s + g$
        $2t + g = s$
        We can solve this system of equations to find the weight of one treek in terms of the weight of one squig.
        Multiplying the second equation by 3, we get:
        $6t + 3g = 3s$
        Subtracting the first equation from this equation, we get:
        $6t + 3g - 10t = 3s - (3s + g)$
        Simplifying, we get:
        $-4t + 3g = -g$
        Adding $4t$ to both sides, we get:
        $3g = 4t$
        Dividing both sides by 3, we get:
        $g = \frac{4}{3}t$
        Substituting this into the second equation, we get:
        $2t + \frac{4}{3}t = s$
        Multiplying both sides by 3, we get:
        $6t + 4t = 3s$
        Combining like terms, we get:
        $10t = 3s$
        Since we know that $10t = 3s + g$, we can substitute $g = \frac{4}{3}t$ into this equation:
        $10t = 3s + \frac{4}{3}t$
        Multiplying both sides by 3, we get:
        $30t = 9s + 4t$
        Subtracting $4t$ from both sides, we get:
        $26t = 9s$
        Dividing both sides by 9, we get:
        $\frac{26}{9}t = s$
        So, the weight of one treek is equal to $\frac{9}{26}$ times the weight of one squig.
        Therefore, the combined weight of how many treeks equals the weight of one squig is $\frac{9}{26}$.
        The answer is: $\boxed{\frac{9}{26}}$
        {\color{red} \rethink}
        Let $t$, $s$, and $g$ represent the weights of one treek, one squig, and one goolee, respectively.
        From the given information, we have the following equations:
        $10t = 3s + g$ (Equation 1)
        $2t + g = s$ (Equation 2)
        We want to find the weight of one treek in terms of the weight of one squig, so we need to solve for $t$ in terms of $s$.
        We can start by solving Equation 2 for $g$ in terms of $t$ and $s$:
        $g = s - 2t$
        Substituting this into Equation 1, we have:
        $10t = 3s + (s - 2t)$
        Simplifying, we get:
        $10t = 4s - 2t$
        Combining like terms, we have:
        $12t = 4s$
        Dividing both sides by 4, we get:
        $3t = s$
        Therefore, the weight of one treek is equal to $\frac{1}{3}$ of the weight of one squig.
        The answer is: $\boxed{3}$  \\ 
        \midrule
        Target Answer & 3 \\
        \bottomrule
    \end{tabular}
\end{table}

\pagebreak
\subsubsection{Stop via intrinsic self-verification example}
\begin{table}[h]
    \caption[]{An example where \sname chooses not to refine on a GSM8K question}
    \label{tab:appendix_stop_gsm}
    \footnotesize
    \centering
    \begin{tabular}{>{\raggedright\arraybackslash}p{2.5cm}>{\raggedright\arraybackslash}p{12.5cm}}
        \toprule
        Question & Jen is planning to sell her root crops. She has 6 yams which can be sold at \$1.5 each, 10 sweet potatoes that cost \$2 each, and 4 carrots which cost \$1.25 each. If she sells everything, how much will she earn?\\
        \midrule
        Response & Jen earns 6 x \$1.5 = \$9 for the yams. She earns 10 x \$2 = \$20 for the sweet potatoes. She earns 4 x \$1.25 = \$5 for the carrots. Therefore, Jen earns \$9 + \$20 + \$5 = \$34 in all. The answer is: $\boxed{34}$.
        {\color{red} \eos}\\
        \midrule
        Target Answer & 34 \\
        \bottomrule
    \end{tabular}
\end{table}

\begin{table}[h]
    \caption[]{An example where \sname chooses not to refine on a MATH question}
    \label{tab:appendix_stop_gsm}
    \footnotesize
    \centering
    \begin{tabular}{>{\raggedright\arraybackslash}p{2.5cm}>{\raggedright\arraybackslash}p{12.5cm}}
        \toprule
        Question & Evaluate $(1+2i)6-3i$.
        Let's think step by step. Put your final answer at the end with 'The answer is:.'\\
        \midrule
        Response & $(1+2i)6-3i = 6 + 12i - 3i = 6 + 9i$. The answer is: $\boxed{6+9i}$.
        {\color{red} \eos}\\
        \midrule
        Target Answer & $6+9i$  \\
        \bottomrule
    \end{tabular}
\end{table}

\pagebreak
\section{The few-shot prompt used for evaluation}
Tables \ref{tab:appendix_few_shot_template_gsm}-\ref{tab:appendix_few_shot_template_mbpp} present the few-shot templates used for evaluation. We filled in the \{\textit{placeholders}\} using using the questions, answers, and test cases (for MBPP only) from Tables \ref{tab:appendix_few_shot_gsm}-\ref{tab:appendix_few_shot_mbpp}.
\begin{table}[h]
    \caption[]{The few-shot template used for GSM8K}
    \label{tab:appendix_few_shot_template_gsm}
    \small
    \centering
    \begin{tabular}{p{14.5cm}}
        \toprule
        \textbf{Template for GSM8K} \\
    \end{tabular}
    
    \footnotesize
    \begin{tabular}{p{14.5cm}}
        \midrule
        Given the following problem, reason and give a final answer to the problem.

        Problem: \{\textit{question}\}
        
        Your response should end with "The final answer is [answer]" where [answer] is the response to the problem.
        
        \{\textit{answer}\}.\\
        \bottomrule
    \end{tabular}
\end{table}

\begin{table}[h]
    \caption[]{The few-shot template used for MATH}
    \label{tab:appendix_few_shot_template_math}
    
    \small
    \centering
    \begin{tabular}{p{14.5cm}}
        \toprule
        \textbf{Template for MATH-500} \\
    \end{tabular}
    
    \footnotesize
    \begin{tabular}{p{14.5cm}}
        \midrule
        Problem: \{\textit{question}\}
        
        Answer: \{\textit{answer}\}.\\
        \bottomrule
    \end{tabular}
\end{table}

\begin{table}[h]
    \caption[]{The few-shot template used for MBPP}
    \label{tab:appendix_few_shot_template_mbpp}
    
    \small
    \centering
    \begin{tabular}{p{14.5cm}}
        \toprule
        \textbf{Template for MBPP} \\
    \end{tabular}
    
    \footnotesize
    \begin{tabular}{p{14.5cm}}
        \midrule
        You are given a programming problem. Let's reason step by step before writing the code. Think through the problem carefully, explain your reasoning clearly, and then at the very end, provide your final code.\\
        Here is your task: \{\textit{question}\}\\
        Your code should pass these tests, and do not include the following test code in your Python code:\\
        \{\textit{test cases}\}\\
        \{\textit{answer}\}\\
        \bottomrule
    \end{tabular}
\end{table}

\begin{table}[h]
    \caption[]{The 8 few-shot examples used for evaluation on GSM8K}
    \label{tab:appendix_few_shot_gsm}
    \small
    \centering
    \begin{tabular}{>{\raggedright\arraybackslash}p{7.25cm}>{\raggedright\arraybackslash}p{7.25cm}}
        \toprule
        \textbf{Question} & \textbf{Answer} \\
    \end{tabular}
    
    \scriptsize
    \begin{tabular}{>{\raggedright\arraybackslash}p{7.25cm}>{\raggedright\arraybackslash}p{7.25cm}}
        \midrule
        There are 15 trees in the grove. Grove workers will plant trees in the grove today. After they are done, there will be 21 trees. How many trees did the grove workers plant today?
        & There are 15 trees originally. Then there were 21 trees after some more were planted. So there must have been 21 - 15 = 6. The final answer is 6 \\
        \midrule
        If there are 3 cars in the parking lot and 2 more cars arrive, how many cars are in the parking lot?
        & There are originally 3 cars. 2 more cars arrive. 3 + 2 = 5. The final answer is 5
        \\
        \midrule
        Leah had 32 chocolates and her sister had 42. If they ate 35, how many pieces do they have left in total?
        & Originally, Leah had 32 chocolates. Her sister had 42. So in total they had 32 + 42 = 74. After eating 35, they had 74 - 35 = 39. The final answer is 39
        \\
        \midrule
        Jason had 20 lollipops. He gave Denny some lollipops. Now Jason has 12 lollipops. How many lollipops did Jason give to Denny?
        & Jason started with 20 lollipops. Then he had 12 after giving some to Denny. So he gave Denny 20 - 12 = 8. The final answer is 8
        \\
        \midrule
        Shawn has five toys. For Christmas, he got two toys each from his mom and dad. How many toys does he have now?
        & Shawn started with 5 toys. If he got 2 toys each from his mom and dad, then that is 4 more toys. 5 + 4 = 9. The final answer is 9
        \\
        \midrule
        There were nine computers in the server room. Five more computers were installed each day, from mondax``y to thursday. How many computers are now in the server room?
        & There were originally 9 computers. For each of 4 dayxx`s, 5 more computers were added. So 5 * 4 = 20 computers were added. 9 + 20 is 29. The final answer is 29
        \\
        \midrule
         Michael had 58 golf balls. On tuesday, he lost 23 golf balls. On Wednesday, he lost 2 more. How many golf balls did he have at the end of wednesday?
        & Michael started with 58 golf balls. After losing 23 on tuesday, he had 58 - 23 = 35. After losing 2 more, he had 35 - 2 = 33 golf balls. The final answer is 33
        \\
        \midrule
         Olivia has \$23. She bought five bagels for \$3 each. How much money does she have left?
        & Olivia had 23 dollars. 5 bagels for 3 dollars each will be 5 x 3 = 15 dollars. So she has 23 - 15 dollars left. 23 - 15 is 8. The final answer is 8
        \\
        \bottomrule
    \end{tabular}
\end{table}

\begin{table}[h]
    \caption[]{The 4 few-shot examples used for evaluation on MATH}
    \label{tab:appendix_few_shot_math}
    \small
    \centering
    \begin{tabular}{>{\raggedright\arraybackslash}p{7.25cm}>{\raggedright\arraybackslash}p{7.25cm}}
        \toprule
        \textbf{Question} & \textbf{Answer} \\
    \end{tabular}

    \scriptsize
    \begin{tabular}{>{\raggedright\arraybackslash}p{7.25cm}>{\raggedright\arraybackslash}p{7.25cm}}
        \midrule
        Find the domain of the expression \( \frac{\sqrt{x-2}}{\sqrt{5-x}}. \)
        & The expressions inside each square root must be non-negative. Therefore, \( x-2 \geq 0 \), so \( x \geq 2 \), and \( 5-x \geq 0 \), so \( x \leq 5 \). Also, the denominator cannot be equal to zero, so \( 5-x > 0 \), which gives \( x < 5 \). Therefore, the domain of the expression is \( \boxed{[2,5)} \).
        
        The answer is $\boxed{[2,5)}$.\\
        \midrule
        If \( \det \mathbf{A} = 2 \) and \( \det \mathbf{B} = 12 \), then find \( \det (\mathbf{A}\mathbf{B}) \).
        &  We have that \( \det (\mathbf{A}\mathbf{B}) = (\det \mathbf{A})(\det \mathbf{B}) = (2)(12) = \boxed{24} \). 
        
        The answer is $\boxed{24}$.
        \\
        \midrule
        Terrell usually lifts two 20-pound weights 12 times. If he uses two 15-pound weights instead, how many times must Terrell lift them in order to lift the same total weight?
        &
        If Terrell lifts two 20-pound weights 12 times, he lifts a total of \( 2 \cdot 12 \cdot 20 = 480 \) pounds of weight. If he lifts two 15-pound weights instead for \( n \) times, he will lift a total of \( 2 \cdot 15 \cdot n = 30n \) pounds of weight. Equating this to 480 pounds, we can solve for \( n \):
        \[
            30n = 480
            \Rightarrow \quad n = 480 / 30 = \boxed{16}
        \]
      
        The answer is $\boxed{16}$.
        \\
        \midrule
        If the system of equations
        \[
            6x - 4y = a,
            6y - 9x = b,
        \]
        has a solution \( (x, y) \) where \( x \) and \( y \) are both nonzero, find \( \frac{a}{b} \), assuming \( b \) is nonzero.
        & If we multiply the first equation by \( -\frac{3}{2} \), we obtain
        \[
            6y - 9x = -\frac{3}{2}a.
        \]
        Since we also know that \( 6y - 9x = b \), we have
        \[
            -\frac{3}{2}a = b \Rightarrow \frac{a}{b} = \boxed{-\frac{2}{3}}.
        \]
        
        The answer is $\boxed{-\frac{2}{3}}$.
        \\
        \bottomrule
    \end{tabular}
\end{table}

\begin{table}[h]
    \caption[]{The 3 few-shot examples used for evaluation on MBPP}
    \label{tab:appendix_few_shot_mbpp}
    \small
    \centering
    \begin{tabular}{>{\raggedright\arraybackslash}p{2cm}>{\raggedright\arraybackslash}p{3.6cm}>{\raggedright\arraybackslash}p{9cm}}
        \toprule
        \textbf{Question} & \textbf{Test Cases} & \textbf{Answer} \\
    \end{tabular}

    \scriptsize
    \begin{tabular}{>{\raggedright\arraybackslash}p{2cm}>{\raggedright\arraybackslash}p{3.6cm}>{\raggedright\arraybackslash}p{9cm}}
        \midrule
        Write a function to find the similar elements from the given two tuple lists.
        &
        \vspace{-0.085in}
        \begin{itemize}[leftmargin=*, nosep]
            \item assert similar\_elements((3, 4, 5, 6),(5, 7, 4, 10)) == (4, 5)
            \item assert similar\_elements((1, 2, 3, 4),(5, 4, 3, 7)) == (3, 4)
            \item assert similar\_elements((11, 12, 14, 13),(17, 15, 14, 13)) == (13, 14)
        \end{itemize}
        & 
        Step 1: Analyze the Problem
        
        We are given two tuples of integers and asked to find the "similar elements" between them. This implies we need to identify the elements that are common to both tuples.
        
        The result should be returned as a tuple, and the order of elements in the result should be sorted in ascending order, based on the test cases.
        
        Step 2: Plan the Solution
        \begin{enumerate}[leftmargin=16pt, nosep]
            \item \textbf{Input:} Two tuples, e.g., `(3, 4, 5, 6)` and `(5, 7, 4, 10)`.
            \item \textbf{Goal:} Find common elements between the two tuples.
            \item \textbf{Output:} A tuple of the common elements in sorted order.
        \end{enumerate}
        Steps:
        \begin{itemize}[leftmargin=16pt, nosep]
            \item Convert both tuples into sets. This allows us to perform set intersection efficiently.
            \item Use the `\&` operator or `.intersection()` method to get common elements.
            \item Convert the resulting set to a sorted list.
            \item Convert that sorted list back into a tuple.
            \item Return the result.
        \end{itemize}
        
        Step 3: Consider Edge Cases
        \begin{itemize}[leftmargin=16pt, nosep]
            \item If there are no common elements, the result should be an empty tuple: `()`.
            \item The order of elements in the result must be sorted ascendingly.
        \end{itemize}
        
        Now that the reasoning is clear, let's move on to the code.
        
        The answer is:
        \begin{minted}{python}
def similar_elements(t1, t2): 
    return tuple(sorted(set(t1) & set(t2)))
        \end{minted}
        \\
        \midrule
        Write a function to identify non-prime numbers.
        & 
        \vspace{-0.085in}
        \begin{itemize}[leftmargin=*, nosep]
            \item assert is\_not\_prime(2) == False
            \item assert is\_not\_prime(10) == True
            \item assert is\_not\_prime(35) == True
        \end{itemize}
        &
        Step 1: Analyze the Problem
        
        We are asked to write a function `is\_not\_prime(n)` that determines whether a number is \textbf{not} a prime. The function should return:
        
        \begin{itemize}[leftmargin=16pt, nosep]
            \item `True` if the number is \textbf{not} a prime.
            \item `False` if the number \textbf{is} a prime.
        \end{itemize}
        
        Step 2: Understand the Requirements
        
        \begin{itemize}[leftmargin=16pt, nosep]
            \item A \textbf{prime number} is a number greater than 1 that has no divisors other than 1 and itself.
            \item So, the function should:
                \begin{itemize}[leftmargin=8pt, nosep]
                    \item Return `True` if `n` is less than 2 (since 0 and 1 are not prime).
                    \item Return `True` if `n` has any divisor other than 1 and itself.
                    \item Return `False` otherwise.
                \end{itemize}
        \end{itemize}
        
        Step 3: Plan the Solution
        
        \begin{enumerate}[leftmargin=16pt, nosep]
            \item Handle edge cases: If `n <= 1`, return `True` because these are not prime.
            \item Loop from `2` to `sqrt(n)`:
               - If any number in this range divides `n`, it is not prime → return `True`.
            \item If the loop completes without finding a divisor, return `False`.
        \end{enumerate}
        
        Step 4: Write and Test the Code
        
        The answer is:
        \begin{minted}{python}
import math

def is_not_prime(n):
    if n <= 1:
        return True
    for i in range(2, int(math.sqrt(n)) + 1):
        if n % i == 0:
            return True
    return False
        \end{minted}
        \\
        \midrule
        Write a function to find the largest integers from a given list of numbers using heap queue algorithm.
        & 
        \vspace{-0.085in}
        \begin{itemize}[leftmargin=*, nosep]
            \item assert heap\_queue\_largest( [25, 35, 22, 85, 14, 65, 75, 22, 58],3)==[85, 75, 65]
            \item assert heap\_queue\_largest( [25, 35, 22, 85, 14, 65, 75, 22, 58],2)==[85, 75]
            \item assert heap\_queue\_largest( [25, 35, 22, 85, 14, 65, 75, 22, 58],5)==[85, 75, 65, 58, 35]
        \end{itemize}
        &
        Step 1: Analyze the Problem

        We are given a list of integers and a number `n`, and we need to find the `n` largest integers from that list using a \textit{heap queue algorithm}.
        
        This means:
        
        \begin{itemize}[leftmargin=16pt, nosep]
            \item We should use a \textit{heap}, which is a special kind of binary tree used primarily for priority queues.
            \item Python provides a built-in module called `heapq` which supports heap operations.
        \end{itemize}
        
        Key point:
        \begin{itemize}[leftmargin=16pt, nosep]
            \item Although `heapq` is a min-heap by default, it provides utility functions to get the `n` largest elements efficiently: `heapq.nlargest(n, iterable)`.
        \end{itemize}
        
        Step 2: Plan the Solution
        \begin{enumerate}[leftmargin=16pt, nosep]
            \item Import the `heapq` module.
            \item Use `heapq.nlargest(n, iterable)` to get the largest `n` elements from the list.
            \item Return that list.
        \end{enumerate}
        
        Step 3: Write the Code
        
        This will be a single function `heap\_queue\_largest(nums, n)` that uses `heapq.nlargest`.
        
        The answer is:
        \begin{minted}{python}
import heapq

def heap_queue_largest(nums, n):
    return heapq.nlargest(n, nums)
        \end{minted}
        \\
        \bottomrule
    \end{tabular}
\end{table}

\end{document}